\documentclass[sigconf]{acmart}
\usepackage[all,poly]{xy}

\usepackage{tikz}
\usepackage{tikz,ifthen}

\usepackage{listings}
\usepackage{xcolor}

\newtheorem{thm}{Theorem}[section]

\newtheorem{lem}{Lemma}
\newtheorem{cor}{Corollary}

\usepackage[capitalise]{cleveref}

\crefname{thm}{Thm.}{}
\crefname{prop}{Prop.}{}
\crefname{lem}{Lem.}{}
\crefname{defn}{Def.}{}
\crefname{exa}{Example}{}
\crefname{exe}{Exercise}{}
\crefname{rem}{Rem.}{}
\crefname{cor}{Cor.}{}
\crefname{prob}{Problem}{}
\crefname{figure}{Fig.}{}
\crefname{table}{Table}{} 

\AtBeginDocument{%
  }

\setcopyright{acmlicensed}
\copyrightyear{2025}
\acmYear{2025}
\acmDOI{XXXXXXX.XXXXXXX}
\acmConference[ISSAC 2025]{International Symposium on
Symbolic and Algebraic Computation 2025}{July 28 - August 1, 2025}{Guanajuato, Mexico}
\acmISBN{978-1-4503-XXXX-X/2018/06}

\def\sig{\mbox{sig} }

\newcommand{\abs}[1]{\left\lvert\mspace{1mu}#1\mspace{1mu}\right\rvert}

\def\Gal{\mbox{Gal }}
\def\wh{\mathfrak H_k} 
\def\GL{GL}
\def\p{\mathfrak p}
\def\sig{\mbox{sig }}
\def\F{\mathbb F}
\def\chara{\mbox{char }}
\def\deg{\mbox{deg }}
\def\d{\delta} 
\def\D{\Delta}
\newcommand{\card}[1]{\left\lvert\mspace{1mu}#1\mspace{1mu}\right\rvert}

\def\SL{\mbox{SL}}
\def\<{\langle}
\def\>{\rangle}

\def\g{\gamma}

\begin{document}
\title{Galois groups of polynomials  and neurosymbolic networks}

\author{Elira Shaska}
\authornote{Both authors contributed equally to this research.}
\affiliation{%
  \institution{Department of Computer Science, \\Oakland University}
  \city{Rochester}
  \country{USA}
  }
\email{elirashaska@oakland.edu}
 
\author{Tony Shaska}
\affiliation{%
  \institution{Department of Mathematics and Statistics, \\Oakland University}
  \city{Rochester}
  \country{USA}
  }
\email{tanush@umich.edu}

\renewcommand{\shortauthors}{Shaska et al.}

\begin{abstract}
This paper introduces a novel approach to understanding Galois theory, one of the foundational areas of algebra, through the lens of machine learning. By analyzing polynomial equations with machine learning techniques, we aim to streamline the process of determining solvability by radicals and explore broader applications within Galois theory. This summary encapsulates the background, methodology, potential applications, and challenges of using data science in Galois theory.

More specifically, we design  a neurosymbolic network to classify Galois groups and show how this is more efficient than usual neural networks.  We discover some very interesting distribution of polynomials for groups not isomorphic to the symmetric groups and alternating groups.  
\end{abstract}



\received{30 January 2025}

\maketitle

\section{Introduction}
Galois theory, a cornerstone of modern algebra, provides profound insights into the solvability of polynomial equations. Since its inception by Évariste Galois, it has explained why there are no general formulas for polynomials of degree five or higher by radicals, unlike the well-known quadratic, cubic, and quartic formulas. 
While traditional methods allow us to determine solvability for lower-degree polynomials through invariants like discriminants, the complexity escalates dramatically for higher degrees, where the Galois group might not be solvable, leading to no radical solution.

This project embarks on a  journey to merge the abstract realm of Galois theory with the practical capabilities of machine learning (ML). Our goal is to harness ML's pattern recognition and prediction abilities to address some of the most challenging aspects of Galois theory, potentially revolutionizing our understanding and approach to polynomial solvability and related problems. 

We propose an approach where we   generate datasets of polynomials with known Galois groups. Key to our approach will be identifying or creating features from polynomials that are indicative of Galois group properties or solvability. These might include traditional invariants of binary forms   or novel features derived from root distributions or algebraic properties. Using supervised learning on neurosymbolic networks, we aim to predict the Galois group or solvability of polynomials.
By learning from simpler polynomials, we hope to generalize these insights to more complex polynomials, possibly using techniques like transfer learning where models adapt knowledge from one task to another.

This integration could lead to automated solvability prediction, offering mathematicians tools to quickly assess if a polynomial can be solved by radicals, and might uncover patterns or invariants not yet recognized by traditional mathematics. The methodology could extend to other areas like field theory or algebraic geometry. However, several challenges loom, including the computational cost of handling high-degree polynomials, ensuring interpretability of ML models to enhance theoretical understanding, and balancing between providing practical tools and contributing to the theoretical body of Galois theory.

This project stands at the intersection of pure mathematics and cutting-edge computational science. By leveraging machine learning, we aim not only to solve practical problems within Galois theory but also to catalyze new theoretical advancements. This exploration could redefine how we approach some of the oldest and most fundamental questions in algebra, potentially opening new avenues for research in both mathematics and computer science.

We discovered that neurosymbolic networks are the most suitable approach for this type of problem.  
A neurosymbolic network is a type of artificial intelligence system that combines the strengths of neural networks (good at pattern recognition) with symbolic reasoning (based on logic and rules) to create models that can both learn from data and reason through complex situations, essentially mimicking human-like cognitive abilities by understanding and manipulating symbols to make decisions. This approach aims to overcome the limitations of either method alone, providing better explainability and adaptability in AI systems. 

The basic foundation of Galois groups of polynomials over $\mathbb{Q}$ are briefly described. We cover in detail the solution of cubics, quartics, and quintics not only to put things in proper context but also to emphasize that each degree is different. There is no universal method in Galois theory that works for every degree, which strongly suggests that AI models should be tailored specifically for each degree. This indicates that neurosymbolic networks might be the best approach for designing models which not only predict the Galois group but also aim to derive solution formulas by radicals (when the group is solvable) and express these formulas in terms of invariants.


We show how to create databases of polynomials, providing a glimpse into how quickly computations can escalate. We detail how we build databases for cubics, quartics, and quintics and uncover some surprising trends even for such small degree polynomials where the theory is well-known. For instance, we find how rare it is for the cyclic group $C_n$ to be the Galois group of a degree $n$ polynomial. For example, among roughly $20^6$ quintic polynomials of height $\leq 10$, only three (up to $\overline{\mathbb{Q}}$-isomorphism) have a Galois group isomorphic to $C_5$, with a total of 20 polynomials (counting twists) corresponding to these three classes. Training an AI model to identify such rare cases might indeed be an impossible task without the use of symbolic methods. Our data could serve various purposes, such as checking Malle's conjecture on Galois groups, verifying results by Bhargava et al. on the number of quartics with bounded heights, or comparing the height of polynomials with the weighted height of invariants.

In section five, we offer a glimpse of what a neurosymbolic network might look like for this application. This is not a fully developed product yet, as it could be refined with many symbolic layers based on theoretical knowledge. However, it shows that for small degrees, it can work relatively well. While there might not be a compelling reason to use AI models to predict the Galois group for degrees $d=3, 4, 5$, this approach could prove to be very successful  for higher degrees.

We hope this paper will encourage mathematicians and computer scientists to explore the use of AI in mathematical research, particularly in tackling classical problems of mathematics. Although this is a modest attempt to incorporate such methods into Galois theory, the rapid development of Artificial Intelligence promises new and innovative applications in mathematics.


\section{Galois groups of  polynomials}
Let $\F$ be a perfect field.  For simplicity we only consider the case when $\chara \F=0$.  
Let $ f(x) $ be a degree $n=\deg f$  irreducible polynomial in $ \F[x]$ which is factored as follows:
\begin{equation} f(x)= (x-\alpha_1) \dots (x-\alpha_n)\end{equation} 
in a splitting field $E_f$. Then, $E_f/\F$ is Galois because is a normal extension and separable. The group  $ \Gal (E_f/\F)$ is called \textbf{the Galois group} of $ f(x)$ over $\F$ and  denoted by $\Gal_\F (f)$. The elements of $\Gal_\F (f)$  permute roots of $ f(x)$. Thus, the Galois group of polynomial has an isomorphic  copy embedded  in $S_n$, determined up to conjugacy by $f$.
%
 
\begin{lem}  The following are true:
\begin{enumerate}
\item   $\deg f \, \mid  \, |G|$ 
\item Let $G=\Gal_\F (f)$ and $H=G\cap A_n$. Then $H= \Gal (E_f / \F(\sqrt{\D_f}))$. In particular, $G$ is contained in the alternating group $A_n$ if and only if  the discriminant $\D_f$ is a square in $\F$.
\item The irreducible factors of $f$ in $\F[x]$ correspond to the orbits of $G$. In particular, $G$ is a transitive subgroup of $S_n$ if and only if  $f$ is irreducible.
\end{enumerate} 
\label{sp_field}
\end{lem}

\proof  The first part is a basic property of the splitting field $E_f$. 
(ii) We have $\D_f=d_f^2$, where $d_f=\prod_{i>j}  (\alpha_i-\alpha_j)$. For $g\in G$ we have $g(d_f)=\mbox{sgn}(g)d_f$. Thus $H=G\cap A_n$ is the stabilizer of $d_f$ in $G$. But this stabilizer equals
$\Gal (E_f /\F (d_f))$. Hence the claim. \\
(iii) $G$ acts transitively on the roots of each irreducible factor of $f$.
\qed

When  $n=2$   then  $f(x) = a_2 x^2 + a_1 x + a_0$. Thus,   $\D_f = a_1^2 -  4a_0 a_2$. Hence $\Gal(f) \cong A_2=\{1\}$ if and only if  $\D_f$ is a square.

\begin{lem}\label{gal-lin-subs}
Let $f(x)\in \F[x]$ be an irreducible polynomial with degree $n$. Then $\Gal_\F (x)$ is an affine invariant of $f(x)$.  Hence, $\Gal (f)\cong\Gal(g)$ for any  $g(x)=f(ax+b)$,     for $a, b \in \F$ and $a\neq 0$.
\end{lem}

\subsection{Cubics}
Let $ f(x)$ be an irreducible cubic in $\F[x]$. Then  $ [E_f:\F] = 3$ or 6.  
Thus,  $\Gal_\F (f) \cong A_3$ if and only if $\D_f$ is a square in $\F$,   otherwise $ \Gal_\F (f) \cong S_3$.
\begin{lem}
Let $f(x) \in \F[x]$ be an irreducible cubic. 
Then  $G=A_3$ if and only if  $\D_f$   is a square in $\F$.     Moreover,  
\begin{enumerate}
\item $\D_f>0$ if and only if  $f$ has three distinct real roots.
\item  $\D_f<0$ iff $f$ has one real root and two non-real complex conjugate roots.
\end{enumerate}  
\end{lem}

Since both $A_3$ and $S_3$ are solvable, we should be able to determine formulas to give the roots of $f(x)$ in terms of radicals. These formulas  are known as Cardano's formulas. 
Hence, for cubics  t we can determine the Galois group simply by conditions on invariants.

\subsection{Quartics}
Let $f(x)\in \F [x]$ be an irreducible quartic. Then $G:=\Gal (f) $ is a transitive subgroup of $S_4$.
Furthermore,  $4 \mid \, \card{G}$, see  \cref{sp_field}. So the order of $G$ is 4, 8, 12, or 24. 
It can be easily checked that   transitive subgroups of $S_4$ of  order   4, 8, 12, or 24  are isomorphic to one of  
\begin{equation}\label{grps:d=4}
C_4, \; D_4, \;  V_4, \; A_4, \; S_4. 
\end{equation}
Any quartic in $\F [x]$ can be normalized as 
\begin{equation}
f(x) = x^4 + ax^2 +   bx +  c    = (x-\alpha_1)  \dots  (x-\alpha_4)      
\end{equation}
with $a,b,c\in \F$. Let $E_f= \F (\alpha_1, \dots,\alpha_4)$ be the splitting field of $f$ over $\F$. Since $f$ has no $x^3$-term, we have $\alpha_1+ \dots +\alpha_4=0$. We assume $\D_f\ne0$, so $\alpha_1, \dots,\alpha_4$ are distinct. Let $G= \Gal_\F (f)$, viewed as a subgroup of $S_4$ via permuting  $\alpha_1, \dots,\alpha_4$.

 There are 3 partitions of $\{1, \dots,4\}$ into two pairs. $S_4$ permutes these 3 partitions, with kernel
\begin{equation}
V_4=\{(12)(34),(13)(24),(14)(23),id\}.
\end{equation}
Thus $S_4/V_4\cong S_3$, the full symmetric group on these 3 partitions. Associate with these partitions the elements
\begin{equation}
\beta_1 = \alpha_1\alpha_2+\alpha_3\alpha_4,   \quad 
 \beta_2 = \alpha_1\alpha_3+\alpha_2\alpha_4,  \quad 
  \beta_3  = \alpha_1\alpha_4+\alpha_2\alpha_3
  \end{equation}
of $E_f$. If $\beta_1=\beta_2$ then $\alpha_1(\alpha_2-\alpha_3)=\alpha_4(\alpha_2-\alpha_3)$, a contradiction. Similarly, $\beta_1, \beta_2, \beta_3$ are 3 distinct elements. Then $G$ acts as a subgroup of $S_4$ on $\alpha_1, \dots,\alpha_4$, and as the corresponding subgroup of $S_3\cong S_4/V_4$ on $\beta_1, \dots,\beta_3$. Thus the subgroup of $G$  fixing all $\beta_i$ is $G \cap V_4$. This proves the following: 
  
\begin{lem} 
The subgroup $G \cap V_4\leq G$ corresponds to the subfield $\F (\beta_1, \beta_2, \beta_3)$, which  is the splitting field over $\F$ of the cubic polynomial (\emph{cubic resolvent})
\begin{equation}
g(x)  =  (x-\beta_1) (x-\beta_2) (x-\beta_3)   =    x^3 - a x^2 -  4cx +  -b^2+4ac    
 \end{equation}
\end{lem}

The roots $\beta_i$ of the cubic resolvent can be found by Cardano's formulas. The extension 
$
\F(\alpha_1, \dots ,\alpha_4)/k(\beta_1, \beta_2, \beta_3)
$ 
has Galois group $\le V_4$, hence is obtained by adjoining at most two square roots to $\F (\beta_1, \beta_2, \beta_3)$. Moreover,   $\D(f,x)=\D(g,x)$.
In general, for   an irreducible quartic
\[
f(x)= x^4 + ax^3 + bx^2 + cx+d
\]
we   can first eliminate the coefficient of $x^3$ by the substituting $x$ with $x - \frac a 4$.  In terms of the binary forms this corresponds to the transformation
$
 (x, y) \to  \left(  x-    \frac a 4 y, y   \right)
$
 and  the new quartic is   $f^M$ for $M=\begin{bmatrix}   1 & - a/4 \\  0 & 1   \end{bmatrix}$.   Since $M \in \SL_2 ({\mathbb Q})$ then $\det M =1$ and the invariants of $f^M$ are the same as those of $f$:    
\begin{equation}\label{inv4}
\begin{split}
\xi_0 (f) & =  2 a_0 a_4 - \frac{a_1 a_3}2 + \frac{a_2^2}{6}      \\
\xi_1 (f) & =  a_0 a_2 a_4 - \frac{3 a_0 a_3^2}{8} - \frac{3 a_1^2 a_4}{8} + \frac{a_1 a_2 a_3}{8} - \frac{a_2^3}{36}   \\
\end{split}
\end{equation}
Moreover     $g(x)$  is
\begin{equation}
g(x) := x^3-b x^2+(a c-4 d) x-a^2 d+4 b d-c^2.
\end{equation}
The discriminant of $f(x)$ is the same as the discriminant of $g(x)$.
We denote by   $d:= [\F(\beta_1, \beta_2, \beta_3) : \F]$.  Then we have the following:

\begin{lem}
The Galois group of $f(x)$ is one of the following:
\begin{enumerate}
\item  $d=1 \Longleftrightarrow G \cong V_4$.
\item $d=3 \Longleftrightarrow G \cong A_4$.
\item  $d=6 \Longleftrightarrow G \cong S_4$.
\item If $d=2$ then we have
\subitem   a)  $f(x)$ is irreducible over $F \Longleftrightarrow G \cong D_4$
\subitem   b)  $f(x)$ is reducible over $F \Longleftrightarrow G \cong C_4$
\end{enumerate}
\end{lem}

\subsubsection{Solving quartics}
The element $(\alpha_1+\alpha_2)(\alpha_3+\alpha_4)$ is fixed by $G\cap V_4$, hence lies in $K(\beta_1, \beta_2, \beta_3)$. We find
\begin{equation}
-(\alpha_1+\alpha_2)^2 = (\alpha_1+\alpha_2)(\alpha_3+\alpha_4) =  \beta_2+\beta_3
\end{equation}
By this and symmetry we get \textbf{Ferrari's formulas}; see \cite{2024-03}.
This completes the case for the quartics. 
 
\subsection{Quintics}
We assume the reader is familiar with some of the classical works in Galois theory \cite{king, solv-sextics, berwick-28, helmut-book, serre}
\begin{lem}\label{quintics}
Let $f(x)\in \F[x]$ be an irreducible quintic. Then its Galois group is one of the following
$C_5$, $ D_5$,       $F_5= AGL(1,5)$,    $A_5$, $S_5$. 
\end{lem}

\proof $G$ is transitive, hence its 5-Sylow subgroup is isomorphic to  $C_5$ (generated by a 5-cycle). If  $C_5$ is not normal, then $G$ has at least 6 of 5-Sylow subgroups; then $\card{G}\ge 6\cdot 5=30$, hence $[S_5:G]\le 4$ which implies $G = S_5, A_5$. If $C_5$ is normal in $G$ then $G$ is conjugate either $C_5$, $D_5$ (dihedral group of order 10) or $F_5= AGL(1,5)$, the full normalizer of $C_5$ in $S_5$, of order 20  (called also the Frobenius group of order 20).
\qed

If the discriminant of the quintic is a square in $\F$ then $\Gal (f) $ is contained in $A_5$. Hence, it is  $C_5, D_5$, or $A_5$.

\subsubsection{Solvable quintics}

If $ G = S_5,  A_5$ then the equation $f(x) = 0$ is not solvable by radicals. We want to investigate here the case $G$ is not isomorphic to $S_5$ or  $A_5$.  Let $f(x)$ be an irreducible quintic in $\F[x]$ given by
\begin{equation}
f(x) =  x^5 + c_4x^4 +   \cdots  +  c_0   =  (x-\alpha_1) \cdots  (x-\alpha_5)    
\end{equation}
Let $G = \Gal (f)$, viewed as a (transitive) subgroup of  $S_5$ via permuting the (distinct) roots $\alpha_1, \cdots,\alpha_5$. As before  $E_f = \F(\alpha_1, \cdots,\alpha_5)$
denotes the splitting field.

A 5-cycle in $S_5=\mbox{Sym}(\{1, \dots,5\})$ corresponds to an oriented pentagon with vertices $1, \dots ,5$. A 5-cycle and its inverse correspond to a (non-oriented) pentagon, and the full $C_5$ corresponds to a pentagon together with its "opposite"; see \cite{2024-03} for a visual illustration.  

Thus $F_5$, the normalizer of $C_5$ in $S_5$, is the subgroup  permuting the pentagon and its opposite. $D_5$ is the subgroup of $F_5$ fixing the pentagon (symmetry group of the pentagon), and $C_5$ is the subgroup of rotations.    For example, $F_5$   is generated by  
\begin{equation}
F_5=\< \sigma, \tau \, \mid \, \sigma^5=\tau^4=(\sigma\tau)^4=\sigma\sigma\tau\sigma^{-1} \tau^{-1} \>, 
\end{equation}
where 
$\sigma=(12345)$ and $\tau = (2453)$.   Thus if $G\le F_5$ then   $G$ fixes

\begin{equation}\label{d_1}
\begin{split}
\d_1  = & \, \, (\alpha_1-\alpha_2)^2 (\alpha_2-\alpha_3)^2 (\alpha_3-\alpha_4)^2 (\alpha_4-\alpha_5)^2 (\alpha_5-\alpha_1)^2 \\
&  \, -  (\alpha_1-\alpha_3)^2 (\alpha_3-\alpha_5)^2 (\alpha_5-\alpha_2)^2 (\alpha_2-\alpha_4)^2 (\alpha_4-\alpha_1)^2\\
\end{split}
\end{equation}
where the first (resp., second) term corresponds to the edges of the pentagon (resp., its opposite).   
There are six 5-Sylow subgroups of $S_5$: 
$H_1  = \<  (1,2,3,4,5)  \> $, 
$H_2  =  \<  (1,2,3,5,4)  \>$, 
$H_3  =  \< (1,2,4,5,3)   \>$, 
$H_4  =   \< (1,2,4,3,5)   \>$, 
$H_5  =   \< (1,2,5,3,4)   \>$, 
$H_6  =   \<  (1,3,4,5,2)  \> $.

To see the full invariance properties, we need to "projectivize" and use the invariants of binary forms. 
Let $y=1=\beta_i$. The generalized version of the $\d_1$'s is $\tilde\d_1$, formed by replacing $\alpha_i-\alpha_j$ by
%
$
D_{ij}  = \det
\begin{bmatrix}
\g_i  & \beta_i \\
\g_j  & \beta_j \\
\end{bmatrix}
$
in the formulas defining the $\d_i$'s. In particular,
\begin{equation}
\tilde\d_1 =  D_{12}^2 D_{23}^2 D_{34}^2 D_{45}^2 D_{51}^2  -  D_{13}^2 D_{35}^2 D_{52}^2 D_{24}^2 D_{41}^2
\end{equation}

Since $S_5$ has six 5-Sylow subgroups let $\d_1, \dots,\d_6$ be the elements associated in this way to the six 5-Sylow's of $S_5$, i.e., to the
six pentagon-opposite pentagon pairs on five given letters; see \cite{2024-03}.

\begin{lem} 
 $\d_i^\sigma = \d_i$ dhe $\d_i^\tau = \d_i$ për $i=1, \dots , 6$.
\end{lem}

Clearly, $G$ permutes $\d_1, \dots,\d_6$. If $G$  is conjugate to a subgroup of $F_5$, it fixes one of $\d_1, \dots,\d_6$; this fixed $\d_i$ must then lie in   $\F$.    

\def\a{\alpha} 

Thus, a necessary condition for the (irreducible) polynomial $f(x)$  to be solvable by radicals is that one  $\d_i$ lies in $\F$, i.e., that the polynomial
\begin{equation}
g(x) =  (x-\d_1)  \cdots (x-\d_6)  \in \F[x]
\end{equation}
has a root in $\F$. It is also sufficient: 

\begin{lem}
If $G$ fixes one $\d_i$ then $G$ is conjugate to a subgroup of $F_5$, provided that $\d_1, \dots,\d_6$ are all distinct. 
\end{lem}

\proof
To check this it is enough to show that     $\d_1, \dots,\d_6$ are mutually distinct (under the hypothesis $\D_f\ne0$).  
hence, we have to show that 
$ \D_f \neq 0 \implies \D_g\neq 0$. 
Using computational algebra we find  $\D_g$  and verify that   
 \[
\D_g   =  \left( (\a_1-\a_2) ( \a_3-\a_4 ) ( \a_4-\a_5 )  ( \a_3-\a_ 5 \right) )^4 \cdot \D_f \cdot  I_2^2  \cdot I_3 \cdot I_4^2 \cdot I_6^2\\
 \]
where  $I_2, I_3, I_4$, and  $I_6$ are given in    \cite{curri}.   Obviously $\D_f\neq 0$ implies that $\a_i - \a_j \neq 0$ for each $i\neq j$. 
This completes the proof. 
\qed

The coefficients of $g(x)$ are symmetric functions in $\alpha_1, \dots,\alpha_5$, hence are polynomial expressions  in $c_0, \dots, c_4$. The goal is to find these expressions explicitly. This gives an explicit criterion to check whether $f(x)=0$ is solvable by radicals.
 
\begin{lem} Let $s_r( x_1, \dots, x_6)$, $r=1, \dots,6$, be the elementary symmetric polynomials
\begin{equation}
s_r = \sum_{i_1<i_2< \dots <i_r} x_{i_1}x_{i_2}  \dots  x_{i_r}.
\end{equation}
Then 
$  d_r: = s_r ( \tilde\d_1, \dots, \tilde\d_6)$
 is a homogeneous polynomial expression in $b_0, \dots,b_5$ of degree $4r$. These polynomials are invariant under the action of $\SL_2(\F)$ on binary
quintics: For any $M\in \SL_2(\F)$ the quintic $f^M$ has the same associated $d_r$'s.
\end{lem}

\proof For $\alpha_j:=\g_j/\beta_j$ we have $\tilde\d_i=  (\beta_1\cdots\beta_5)^4  \d_i =  b_5^4 \d_i$. Thus 
 $d_r = b_5^{4r} \,  s_r(\d_1,, \dots, \d_6)$. But the $ s_r(\d_1,, \dots, \d_6)$ are polynomial expressions in the $c_j=b_j/b_5$, for   $j=0, \dots,4$. 
 Thus $d_r$ is a rational function  in $b_0, \dots,b_5$, where the denominator is a power of $b_5$. Switching the roles of $x$ and $y$ yields that the denominator is also a power of $b_0$. Thus it is constant, i.e., $d_r$ is a polynomial in $b_0, \dots,b_5$. If we  replace each $\beta_j$ by $c\beta_j$ for a scalar $\lambda$ then each $\tilde\d_i$ gets multiplied by $\lambda^4$, so $d_r$ gets multiplied by $\lambda^{4r}$. Thus $d_r$ is homogeneous of degree $4r$. The rest of the claim is clear.
\qed

There are four basic invariants of quintics, denoted by $J_4, J_8, J_{12},  J_{18}$, of degrees 4,8,12 and 18, such that every $\SL(2,\F)$-invariant polynomial in $b_0, \dots,b_5$ is a polynomial in $J_4,J_8, J_{12}, J_{18}$; see \cite{schur}  or \cite{2024-03}.

By using special quintics one gets linear equations for the coefficients expressing the $d_r$'s in terms of $J_4, J_8, J_{12}$. The result is
due to Berwick; see \cite{king}.
\[
\begin{split}
d_1  & =   -10 J_4  			\\  
d_2  & =   35 J_4^2 + 10  J_8 	\\  
d_3  &=   -60 J_4^3 - 30 J_4 J_8 - 10  J_{12}	 \\  
d_4 & =  55 J_4^4 + 30 J_4^2 J_8 + 25  J_8^2 + 50 J_4 J_{12}\\ 
d_5 & =   -26 J_4^5 - 10 J_4^3 J_8 - 44 J_4 J_8^2 - 59 J_4^2 J_{12}  - 14  J_8 J_{12}\\
d_6 & =  5J_4^6 + 20 J_4^2 J_8^2 + 20 J_4^3 J_{12} + 20 J_4 J_8 J_{12} + 25  J_{12}^2 \\
\end{split}
\]


\begin{lem}\label{gal_5}
Let $f(x)$ be a irreducible quintic over $\F$ and $d_1, \dots, d_6$ defined in terms of the coefficients of $f(x)$ as above. Then $f(x)$ is solvable by radicals if and only if   $g(x)= x^6 + d_1 x^5 + \cdots d_5 x + d_6$
has a root in $\F$.
\end{lem}

\section{Higher degree polynomials}
Next we want to compile some general rules for computing the Galois group of a degree $n$ irreducible polynomial.  We will focus mostly on transitive subgroups of the symmetric group, which provide the candidates for the Galois groups, and the signature of each group which in most cases will determine the group. 

\subsection{Transitive groups}\label{trans-subs}
From the previous discussion we know that  if $f(x) $ is a degree $n$ irreducible polynomial then its Galois group $\Gal (f)$ is a transitive subgroup of $S_n$. 
Using computational group theory and GAP, we can compute list of transitive subgroups for relatively large $n$.  These precompiled lists for every $n$ will be our candidates for Galois groups.
Here is the number of transitive subgroups for $n\leq 47$
\begin{table}[h!]
\centering
\begin{tabular}{|c|c|c|c|c|c|c|c|}
\hline
$n$ & \# Subs & $n$ & \# Subs & $n$ & \# Subs & $n$ & \# Subs \\ 
\hline
5   & 5           & 6   & 16          & 7   & 7           & 8   & 50          \\
9   & 34          & 10  & 45          & 11  & 8           & 12  & 301         \\
13  & 9           & 14  & 63          & 15  & 104         & 16  & 1954        \\
17  & 10          & 18  & 983         & 19  & 8           & 20  & 1117        \\
21  & 164         & 22  & 59          & 23  & 7           & 24  & 25000       \\
25  & 211         & 26  & 96          & 27  & 2392        & 28  & 1854        \\
29  & 8           & 30  & 5712        & 31  & 12          & 33  & 162         \\
34  & 115         & 35  & 407         & 36  & 121279      & 37  & 11          \\
38  & 76          & 39  & 306         & 40  & 315842      & 41  & 10          \\
42  & 9491        & 43  & 10          & 44  & 2113        & 45  & 10923       \\
\hline
\end{tabular}
\smallskip
\caption{Number of transitive subgroups of \( S_n \)}
\end{table}

\noindent Below we list all possible transitive subgroups for $n\leq 19$.   
\begin{table}[h!]
    \centering
    \caption{Transitive Subgroups of \( S_n \) for \( n = 5,  7, 11, 13, 17, 19 \)}
    \begin{tabular}{@{}ll@{}}
        \toprule
        \( n \) & Subgroups \\ 
        \midrule
        5 & $ C_5, D_5, F(5) = 5:4, A_5, S_5 $  \\ 
        7 & $ C_7, D_7, F_{21}(7) = 7:3, F_{42}(7) = 7:6, L(7) = L(3,2), A_7, S_7 $\\ 
        11 & $ C_{11}, D_{11}, F_{55}(11) = 11:5, F_{110}(11) = 11:10, L(11)  $\\
              & $  M_{11}, A_{11}, S_{11} $ \\ 
        13 & $ C_{13}, D_{13}, F_{39}(13) = 13:3, F_{52}(13) = 13:4, F_{78}(13) = 13:6, $ \\
           & $  F_{156}(13) = 13:12,  L(13), A_{13}, S_{13} $ \\ 
        17 & $ C_{17}, D_{17}, F_{68}(17) = 17:4, F_{136}(17) = 17:8, F_{272}(17) = 17:16, $ \\
        & $ L(17), L(17):2 = \text{PZL}(2,16), L(17):4 = \text{PYL}(2,16), A_{17}, S_{17} $ \\ 
        19 & $ C_{19}, D_{19}, F_{57}(19) = 19:3, F_{114}(19) = 19:6, F_{171}(19) = 19:9, $ \\
        & $F_{342}(19) = 19:18, A_{19}, S_{19} $ \\ 
        \bottomrule
    \end{tabular}
\end{table}
As one can see from the above table  the notation used for groups is GAP notation and not suitable for use in Python lists.  
To avoid confusion, in our databases we use either the  GAP Identity or create our own notation tailored to each specific degree. 
\subsection{Reduction modulo $p$}
The reduction method   uses the fact that   every polynomial with rational coefficients can be transformed into a monic polynomial with integer coefficients without changing the  splitting field.
Let $f(x) \in {\mathbb Q}[x]$ be given by
\begin{equation}
f(x)= x^n + a_{n-1} x^{n-1} + \cdots + a_1 x + a_0
\end{equation}
Let $d$ be the common denominator   of all coefficients $a_0, \cdots, a_{n-1}$. Then $g(x):=d \cdot f(\frac x d)$ is a monic polynomial with integer coefficients. Clearly the splitting field of $f(x) $ is the same as the splitting field of $g(x)$. Thus, without loss of generality we can assume that $f(x) \in {\mathbb Z}[x]$ is a monic polynomial
with integer coefficients.

\begin{thm}\textbf{(Dedekind)} 
Let $f(x) \in {\mathbb Z}[x]$ be a monic polynomial such that  $\deg f = n$, $\Gal_{\mathbb Q} (f) = G$, and $p$ a prime such that $p \nmid {\Delta}_f$. If $f_p:=f(x) \mod p \, \, $ factors in ${\mathbb Z}_p [x]$ as a product of irreducible factors of degree     $n_1, n_2, n_3, \cdots, n_k$, 
then $G$ contains a permutation of type     $(n_1)\, (n_2) \, \cdots \, (n_k)$
\end{thm}


The Dedekind theorem can be used to determine the Galois group in many cases since the \emph{type} of permutation in $S_n$ determines the conjugacy class in $S_n$.   Consider for example polynomials of degree 5. The cycle types for all groups that occur as Galois groups of quintics are   easily determined. 

%
\[
\begin{tabular}{|c|c|c|c|}
\hline
\# & Gr        & Id  & signature \\
\hline
1& $C_5$&[5, 1] 	& 	$ [5] $  					\\
2& $D_5$&[5, 2] 	& 	$ [(2)^2, 5]$  				\\
3& $F_5$&[5, 3] 	& 	$ [4 , (2)^2 ,  5]$  			\\
4& $A_5$&[5, 4] 	& 	[   $(2)^2$ , 3 , 5]   			\\
5& $S_5$&[5, 5] 	& 	$[2, (2)^2,  3 , 2\cdot 3 , 4, 5 ]$  \\
\hline
\end{tabular}
\]
Below is the inclusion among the subgroups which we will use to define a symbolic layer for our network. 
 \begin{figure}[h!] 
   \centering
   \includegraphics[width=1.2in]{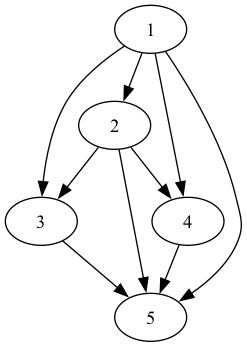} 
   \caption{Lattice of transitive subgroups of $S_5$}
   \label{fig:s5}
\end{figure}

The \emph{signature} of $G$ is the set of such types of permutations.  Notice that for quintics not two subgroups have the same signature. Unfortunately this is not the case for higher degree, so the signature does not always uniquely determines the group. 

\subsection{Polynomials with non-real roots}\label{real-roots}
Let $f(x)\in {\mathbb Q}[x] $ be an irreducible polynomial of  degree $n > 5$.  Denote by $r$ the number of non-real roots of $f(x)$. Since the complex conjugation permutes the roots then $r$ is even, say $r=2s$. By a reordering of the roots we may assume that if $f(x)$ has $r$ non-real roots then
\begin{equation}
\alpha   :=(1, 2) (3, 4)\cdots (r-1, r) \in \Gal (f).
\end{equation}
Since determining the number of non-real roots can be very fast, we would like to know to what extent the number of non-real roots of $f(x)$ determines $\Gal (f)$. The complex conjugation assures that $m(G) \leq r$. The existence of $\alpha$ can narrow down the list of candidates for $\Gal (f)$. However, it is unlikely that the group can be determined only on this information unless $p$ is "large" enough.  In this case the number of non-real roots of $f(x)$ can almost determine the  Galois group of $f(x)$, as we will see in the next section. Nevertheless, the test is worth running for all $p$  since it is very fast and improves the algorithm overall.

 $\Gal (f)$ is determined  for  $\deg (f)$ prime     with $r$ non-real  roots when the degree   is large enough with respect to $r$; see \cite{2004-1}

\begin{thm}
Let $f(x)\in {\mathbb Q}[x] $ be an irreducible polynomial of prime degree $p \geq 5$ and  $r=2s$ be the
number of non-real roots of $f(x)$. If $s$ satisfies     $s \, ( s \log s + 2 \log s + 3) \leq p$,     then $\Gal (f)= A_p, S_p$.
\end{thm}

For a fixed $p$ the above bound is not sharp as we will see below. However, the above theorem can be used successfully if $s$ is fixed. We denote the above bound on $p$ by
\begin{equation}N(r) := \left[ s \, ( s \log s + 2 \log s + 3) \right] \end{equation}
for $r=2s$. Hence, for a fixed number of non-real roots, for  $p \geq N(r)$ the Galois group is always $A_p$ or $S_p$.

\begin{cor} 
Let a polynomial of prime degree $p$ have $r$ non-real roots.   Then $\Gal (f) = A_p$ or $S_p$ if one of the following  holds:
\begin{enumerate} 
\item $r=4$ and $p > 7$,
\item $r=6$ and $p > 13$,
\item $r=8$ and $p > 23$,
\item $r=10$ and $p > 37$,
\end{enumerate}
\end{cor}
The above results can be generalized to every degree, but the result is more technical to be stated here. 

\section{Databases of  polynomials}
In this section we want to create a database of irreducible polynomials $f\in \mathbb Z[x]$ of degree $\deg f =n$.  Data will be stored in a Python dictionary.  A polynomial $f(x) = \sum_{i=0}^n a_i x^i$ will be represented by its corresponding binary form $f(x, y) = \sum_{i=0}^n a_i x^i y^{n-i}$. Hence our points will be points in the projective space ${\mathbb P}^n$, i.e. points with integer coordinates
$
{\mathfrak p} = [a_n : \cdots : a_0] \in {\mathbb P}^n, 
$
such that $\gcd (a_0, \ldots , a_n)$.    Since $f(x)$ is irreducible over ${\mathbb Q}$ and of degree $\deg f=n$, then $a_n\neq 0$ and $a_0\neq 0$.  Moreover, the discriminant ${\Delta}_f\neq 0$. 

%
Let us now trying to generate a dataset with a bounded height $h$ as defined in \cite{2024-03}. 
 We will denote the set of such polynomials by $\mathcal P_n^h$. In other words
\[
\mathcal P_n^h := \left\{  
 \p=[a_n : \cdots : a_0] \in {\mathbb P}^n \; \mid \, a_0 a_n \neq 0,   {\Delta}_f \neq 0,  H_{\mathbb Q} (  \p  )   \leq h
\right\}
\]
where $H_Q$ is defined as in \cite{2024-03}.   
To ensure that the points in the database are not repeated we key the dictionary by the tuples 
$\left(   a_0, \ldots , a_n  \right)$.     A dictionary in Python does not allow key duplicates, which ensures that there are no duplicates in our data. 
For  given $h, n$ the cardinality of $\mathcal P_n^h $ is bounded by 
\[
\# \mathcal P_n^h \leq 4 h^2 (2h+1)^{n-2}
\]
The proof is a straightforward counting argument.   
For a degree $d\geq 3$ and height $h$ one can use \textsl{Sagemath} and creates such sets as:

\begin{verbatim}
PP = ProjectiveSpace(d, QQ)
rational_points = PP.rational_points(h)
\end{verbatim}

We then \emph{normalize} the data by clearing denominators. 
Hence, all our data has integer coordinates. 
Furthermore, we keep only those polynomials which are irreducible over ${\mathbb Q}$. 
For every point ${\mathfrak p} =  [a_n : \cdots : a_0]$ we will compute the following attributes  
\[
\left(   a_0, \ldots , a_n  \right) : 
\left[    
H(f), 
[ \xi_0, \ldots , \xi_n,   {\Delta}_f ],
  \wh({\mathfrak p}),               
 \text{sig},      
     \Gal_{\mathbb Q} (f),       
  \right]
\]
where 
$H(f)$ is   		the height of $f(x)$, 
$[ \xi_0, \ldots , \xi_n ]$ 	generators of the ring of invariants of binary forms of degree $n$ and the discriminant. 
${\Delta}_f$,  
$\wh({\mathfrak p})$ the    		eeighted moduli height  
\text{sig}		the signature, and 
$\Gal_{\mathbb Q} (f)$ the 		 Gap Identity of the Galois group. 

Some of the datasets differ for different degrees. For example for quartics, we also compute the invariants $T$ and $S$ as defined in \cite{dolgachev, 2024-05}
and  the $j$-invariant.  For sextics we compute absolute invariants $t_1, t_2, t_3$; see \cite{2024-03} for details. 
We give a slice of the corresponding dictionary for each $d=3, 4, 5$ which we discuss in the rest of this paper and make all datasets available at \cite{galois-web}.


\subsection{Cubics}
As a simple first exercise we start with irreducible cubics.     We create a database of all rational points $[c_0 : c_1: c_2 :c_3]$  in ${\mathbb P}^3$ with projective height $h\leq 20$ such that 
\[
f(x)= c_0 + c_1 x + c_2 x^2 + c_3 x^3
\]
is an irreducible polynomial in ${\mathbb Q} [x]$.    Since training a model for determining $\Gal (f)$ is trivial in this case we will focus mostly on comparing the naive height with the weighted moduli height and determining how the occurrence of $A_3$ happens with the increase of $h$.

A slice of five random elements of our Python dictionary looks like:

\begin{center}
\begin{tabular}{|c|c|}
\hline
Key & Value \\
\hline
(-1, -9, -20, 1) & [20, 98, 3.1463462836, 'A3'] \\
\hline
(20, -9, -20, 1) & [20, 1458632, 34.752530588, 'A3'] \\
\hline
\hline
\end{tabular}
\end{center}
where the 'key' has the coefficients of the cubic and the entries in 'values' are respectively: naive height, $J_4$ invariant, weighted heigh, and the Galois group. 

\begin{lem}
The total number of rational points of heights in (0 , 20] is = 1 299 200.     From those there are 1 178 856 irreducible polynomials and only     1328 of them have Galois group $C_3$.  Moreover, the distribution of polynomials with Galois group $C_3$ with respect to their naive height is given in \cref{fig:deg3}. 
 \end{lem}

In \cite{2014-1} we give an estimate on the ratio of the moduli height over the naive height for binary sextics.  Such bounds can be given for every degree $d$ polynomial.   In out case of cubics the minimum  ratio is  0.074  for polynomial    
\[
f(x)= 7x^3-5x^2-16x+7\]
 and the 
maximum ratio is  2.008 for 
\[
f(x)=   13x^3-19x^2-20x+13.
\]
  
\begin{lem}
There are only 40 cubics in the database with   height $\leq 5$ and Galois group of order 3. The discriminant ${\Delta}_f$ of those forty polynomials has values ${\Delta}_f=7^2, 3^4, 13^2, 19^2, 31^2$, and $61^2$; see    \cref{tab:deg-3}
\end{lem}

Distribution of points   versus the invariants   is given in \cref{fig:deg3-2}.


\subsection{Quartics}\label{quartics-2}
All quartics are  rational points $[c_0 : c_1: c_2 :c_3:c_4]$  in ${\mathbb P}^3$ with projective height $h\leq 20$ such that 
\[
f(x)= c_0 + c_1 x + c_2 x^2 + c_3 x^3 +c_4 x^4
\]
is an irreducible polynomial in ${\mathbb Q} [x]$.    Other than $S_4$ the other possible Galois groups are $C_4$, $D_4$, $V_4$, and $A_4$.
  We refer to \cref{inv4} for its invariants. However, to avoid denominators we define
\[
J_2= 36 \cdot \xi_0, \qquad J_3= 216\cdot \xi_1, \qquad J_6={\Delta} (f, x)
\]
One can verify that    $J_6 = \frac 1 {27} (4 J_2^3-  J_3^2)$. 
Notice that since   $J_6\neq 0$  we can also define the $\GL_2 ({\mathbb Q})$-invariant or    \emph{$j$-invariant}
\[
j = \frac {J_2^3}  {4 J_2^3 - J_3^2}
\]
A slice of the database for quartics looks as follows:
\begin{table}[h!]
\begin{tabular}{|c|c|}
\hline
Key & Value \\
\hline
(1, -2, -2, -2, 1) & [2, [4, -416], 4.5162, 'D(4)', -6400, -1/2700] \\
\hline
(-1, 2, -1, -2, 1) & [2, [1, 110], 3.23853, 'D(4)', -448, -1/12096] \\
\hline
\end{tabular}
\smallskip
\label{tab:deg-4}
\caption{A slice of the database for quartics}
\end{table}

The increase of the number of polynomials with respect to height seems very comparable to degree 3 and 4.  We present this graphically in \cref{fig:deg4}.

In \cite{2014-1} we give an estimate on the ratio of the moduli height over the naive height for binary sextics.  Such bounds can be given for every degree $d$ polynomial.   In the case of quartics the minimum  ratio is  
0.2236  for the polynomial    
\[
f(x)= x^4 -5x^3 +10x^2-10x+5
\]
and the maximum ratio is  
3.3959 for 
\[
f(x)=   x^4-x^3-x^2-x+1.
\]    
The first quartic has Galois group $C_4$ and the second $F_5$.  We present the ration of the weighted height over the naive height in \cref{fig:deg4-2}

There are 5676 irreducible quartics of naive height $h\leq 10$ with Galois group not isomorphic to $S_4$.  From those 
$D_4$:    5162 polynomials,  $A_4$: 184 polynomials,  $V_4$: 222 polynomials, and $C_4$: 108 polynomials.   In \cref{fig:deg4} we display how the number of such polynomials grows according to the height. 
The 5676 irreducible quartics are up to $\mathbb Z$-equivalence. However, there are only 1231  irreducible quartics up to ${\mathbb Q}$-equivalence, counted by their $j$-invariant.  

In \cite{bhargava-2015}, being  unaware of the weighted height,  the authors define the height of a binary quartic as 
\[
h (f) = \max\{ \abs{J_2}^3, \abs{J_3}^2   \}
\]
Of course this is what we have called the \emph{moduli height} and it is simply the six power  $\wh(f)^6$ of the weighted height.  One of the problems considered in \cite{bhargava-2015} is the number of binary quadratic with bounded height.
%
The authors give necessary and sufficient conditions for $(J_2, J_3)$ to be invariants of an integral quartic.  We verify such conditions in our database. 

The case of quartics is very interesting in its own due to many connections to number theory and elliptic curves and will be the focus of a more detailed investigation in a later stage. 
\subsection{Quintics}
Next we consider the irreducible quintics over ${\mathbb Q}$.  Again polynomial will be identified with points $[c_0: c_1: c_2 : c_3: c_4: c_5]$  in ${\mathbb P}^4$.     The Galois group of an irreducible quintic is   one of the following
$C_5$, $ D_5$,       $F_5= AGL(1,5)$,    $A_5$, $S_5$.   From \cite{2024-05}  the invariants are $\xi_0, \xi_1, \xi_2$ of order 4, 8, 12  respectively.    The expressions of such invariants   suggest we use instead   $J_4 =   - \frac {625} 2 \cdot \xi_0$  and $ J_8  =    1562500\cdot  \xi_1$. 
There are two other invariants $J_{12}$ and $J_{18}$   and there is a degree 36 homogenous polynomial $F(J_4, J_8, J_{12}, J_{18})=0$.   This is a homogenous polynomial of degree 36 in terms of coefficients.  Hence, a degree 2 polynomial in $J_{18}$.    According to Dolgachev \cite[pg. 152]{dolgachev} the discriminant of the quintic is ${\Delta}= J_4^2-128J_8$. 
A  slice of the dictionary for quintics is:

\begin{table}[h!]
\begin{tabular}{|c|c|}
\hline
Key & Value \\
\hline
(-2,-1,0,-2,-2,1) & [2,[-3264,-8152576,-29726998528],7.55,$G_3$] \\
\hline
(1,0,-1,2,-2,1) & [2,[-539,3599,116197],4.81,$G_2$ ] \\
\hline
\end{tabular}
\end{table}

The increase of the number of polynomials with respect to height seems very comparable to degree 3 and 4 as it can be seen in \cref{fig:deg5}.

In \cite{2014-1} we give an estimate on the ratio of the moduli height over the naive height for binary sextics.  Such bounds can be given for every degree $d$ polynomial.   In the case of quintics the minimum 
 ratio is  0.5353  for the polynomial    
 \[
 f(x)= x^5-5x^4+9x^3-9x^2    + 4x -1
 \]
  and the 
maximum ratio is  3.7792 for 
\[
f(x)=   x^5-2x^4-2x^3-x-2.
\]    
The first quintic has Galois group $D_5$ and the second $F_5$.  We present the ration of the weighted height over the naive height in \cref{fig:deg5-2}

\begin{lem}  From all irreducible quintics in $\mathbb Z[x]$ with height $\leq 10$  there are exactly 
20 of them with Galois group $C_5$, 480 with group $F_5$, 900 with group $D_5$, and 1146 with group $A_5$.  Moreover, all polynomials $f$ with   $\Gal (f) \equiv C_5$ and their invariants are listed in \cref{tab:deg-5-$C_5$}.
\end{lem}

\begin{small}
\begin{table}[h!]
\begin{tabular}{|c|c|c|c|c|}
\hline
Key & h & p  & wh    \\
\hline
-1,1,4,-3,-3,1 & 4 &  [4235,4026275,-16076916075]&  8.06  \\
-1,3,3,-4,-1,1 & 4  &   [4235,4026275,-16076916075]&  8.06  \\
1,3,-3,-4,1,1 & 4  &   [4235,4026275,-16076916075]&  8.06  \\
1,1,-4,-3,3,1 & 4  &   [4235,4026275,-16076916075]&  8.06  \\
-1,-2,5,2,-4,1 & 5  &   [4235,4026275,-16076916075]&  8.06  \\
1,4,2,-5,-2,1 & 5  &   [4235,4026275,-16076916075]&  8.06  \\
-1,4,-2,-5,2,1 & 5  &   [4235,4026275,-16076916075]&  8.06  \\
1,-2,-5,2,4,1 & 5  &   [4235,4026275,-16076916075]&  8.06  \\
1,-6,10,-1,-6,1 & 10  &   [4235,4026275,-16076916075]&  8.06  \\
1,-6,-1,10,-6,1 & 10  &   [4235,4026275,-16076916075]&  8.06  \\
-1,-6,-10,-1,6,1 & 10  &   [4235,4026275,-16076916075]&  8.06  \\
-1,-6,1,10,6,1 & 10  &   [4235,4026275,-16076916075]&  8.06  \\
\hline
-1,4,9,-5,-9,1 & 9  &  [113377,2971552001,-47471703427379] &  18.34  \\
-1,9,5,-9,-4,1 & 9  &   [113377,2971552001,-47471703427379] &  18.34  \\
1,9,-5,-9,4,1 & 9  &   [113377,2971552001,-47471703427379] &  18.34  \\
1,4,-9,-5,9,1 & 9  &   [113377,2971552001,-47471703427379] &  18.34  \\
\hline
-1,0,10,5,-10,1 & 10  &   [109375,2392578125,-96893310546875]&  18.18  \\
-1,10,-5,-10,0,1 & 10  &   [109375,2392578125,-96893310546875]&  18.18  \\
1,10,5,-10,0,1 & 10  &   [109375,2392578125,-96893310546875]&  18.18  \\
1,0,-10,5,10,1 & 10  &   [109375,2392578125,-96893310546875]&  18.18  \\
\hline
\end{tabular}
\smallskip
\caption{Quintics of height $\leq 10$ and   $\Gal (f) \cong C_5$}
\label{tab:deg-5-$C_5$}
\end{table}
\end{small}


Data in \cref{tab:deg-5-$C_5$} shows some very interesting trends.  First, There are really only 3 quintics with Galois group $C_5$ up to  $\bar {\mathbb Q}$-isomorphism  since they obviously have the same invariants.  This once more stresses the point that the absolute invariants are really the most effective way of dealing with such databases since they considerable decrease the size of the database.  Furthermore, by decreasing redundancy the learning process of any AI model becomes more efficient.  Some of these issues are further illustrated and discussed in \cite{2024-03}.

Second, the polynomials in  \cite{2024-07} provide interesting     examples of how the height of the binary form can change even for polynomials of such small height.  These are very interesting examples in reduction theory; see \cite{2020-1} and more recently \cite{2024-06}

Finally, the above data emphasizes how rare such cases are.  There are roughly $20^6$ quintic polynomials of height $\leq 10$ and from those only three (up to $\bar {\mathbb Q}$-isomorphism) have Galois group isomorphic to $C_5$.  
Training an AI model to pick such very rare cases might be an impossible task indeed.  We will explore that in the next section. 

\section{neurosymbolic networks}

A neurosymbolic network is a type of artificial intelligence system that combines the strengths of neural networks (good at pattern recognition) with symbolic reasoning (based on logic and rules) to create models that can both learn from data and reason through complex situations, essentially mimicking human-like cognitive abilities by understanding and manipulating symbols to make decisions; this approach aims to overcome limitations of either method alone, providing better explainability and adaptability in AI systems.   They seem to be the most reasonable choice for our approach since we can use all the theoretical knowledge that we have about polynomials and their Galois groups and somehow incorporate this into some machine learning model.   
The area of research on deep learning for symbolic mathematics is very active and has had a lot of activity in the  last few years;  \cite{lample-charton, nsn-2,   england,  england-25}

\noindent  \textbf{Precomputed data for every degree $d$}
For each degree $d$ we precompute two lists:  

i)  " d-grps":  list of transitive subgroups of $S_d$; see \cref{trans-subs}

ii) "d-sig":   list of the signature for every group in " d-grps"

\noindent Such data can be computed using GAP and  group theory.

\noindent  \textbf{Layers:}
Next we describe three symbolic layers that we implement in our model. 


\noindent \textbf{Real roots layer:}
If the polynomial has enough real roots then from \cref{real-roots} the group is $A_d$ or $S_d$.  Computing the real roots is usually easy since it can be done with numerical methods.  Hence, for high enough degree $d$ it is usually an efficient method to compute the number of the real roots of $f(x)$.

The algorithm for finding the number of real roots of a polynomial using Sturm's theorem involves constructing a Sturm sequence, which starts with the polynomial 
$f(x)$   and its derivative, followed by successive remainders from polynomial division, with signs reversed. The number of real roots in a given interval is determined by evaluating the sequence at the interval endpoints and counting sign changes in the resulting values. By substituting large finite values 
$(\pm 10^{10})$  for infinity, the method can approximate the count of real roots over the entire real line. This approach works efficiently for polynomials with integer or rational coefficients.   Its implementation is shown in \cref{app:quintics}. 

\noindent \textbf{Signature layer: }
The first symbolic reasoning layer that we apply to our data is the \emph{signature layer}.  This layer for every point $key = (a_0, \ldots , a_d)$ creates the polynomial $f(x)$ and computes the factorization $f_p (x)$ for a list of primes $p$.  Normally we use $p =2, 3, 5, 7$.  This signature $\sig (key)$ is compared with the list of possible signatures for the degree $d$.  The field of \emph{groups} for this entry is updated with the list of all groups which admit this signature.  If length of 
$L[key][groups]=1$ then  $\Gal (f)$ is uniquely determined and the training is done.    A Python implementation is shown in \cref{app:quintics}. 

\noindent \textbf{Discriminant layer:}
The discriminant is computed for all polynomials in the precomputed data stage, but it is not factored.  This layer is activated only if the entry has as Galois group candidates which are contained or not in the alternating group  $A_d$.   Since this layer can slow down considerably the model, we only activate it as a last resort.

\noindent \textbf{Implementation and efficiency}
 We implement this approach and test it for quartics and quintics databases that we created for this paper. The case of cubics is quite trivial from the point of view of Galois theory and we ignore it here.  While both quartics and quintics are well understood and we don't need any AI model to find out the Galois group, they do provide nice test cases which can tell us how reasonable and efficient such approach is.  We study sextics in more detail in \cite{2024-07}. 

\noindent \textbf{Galois Network:}
We design a network that integrates numerical learning with symbolic reasoning to classify polynomials based on their Galois group properties. The core of this system, which we call the GaloisNetwork, processes polynomial coefficients and leverages mathematical insights to predict the corresponding group labels. This hybrid approach combines the power of deep learning with domain-specific rules, ensuring both accuracy and interpretability.

The input to the network consists of feature vectors derived from polynomial coefficients. We compute these features using mathematical invariants, such as root counts and other Galois group characteristics, creating a robust representation of each polynomial. The features are standardized to improve model performance and are then split into training and validation datasets. Labels representing Galois groups are mapped to numeric values for compatibility with the learning process.

The GaloisNetwork itself is a fully connected feedforward neural network. It begins with an input layer that matches the size of the feature vectors. The network includes three hidden layers, each with 64 neurons and ReLU activation functions, providing the capacity to learn complex patterns in the data. Finally, an output layer produces a probability distribution over all possible Galois group labels using a softmax activation. This architecture allows the network to effectively capture the relationships between features and group classifications.

Training the network involves minimizing a cross-entropy loss function using the Adam optimizer. Over 100 epochs, the network iteratively updates its weights through backpropagation, ensuring that it learns to align its predictions with the true labels. To monitor its progress, we periodically evaluate the model on a validation set, tracking the loss and refining the learning process.

To enhance the model's predictions, we implement a post-processing step that applies domain-specific rules. For example, if the number of real roots of a polynomial exceeds a certain threshold, the prediction is adjusted to align with known Galois group properties. This rule-based layer ensures that the network respects established mathematical principles, making its outputs both reliable and interpretable.

Finally, we evaluate the system using accuracy metrics, confusion matrices, and detailed classification reports. These evaluations demonstrate the effectiveness of combining numerical learning with symbolic reasoning. By integrating these two paradigms, our design not only achieves high accuracy but also maintains alignment with the underlying mathematical structure of the problem, providing a powerful tool for analyzing polynomials through the lens of their Galois groups.
Details of the implementation and links to databases are provided in \cite{2024-05} and 
\href{https://www.risat.org/galois.html}{https://www.risat.org/galois.html}; see \cite{galois-web}.

\section{Concluding remarks}
This paper introduces an innovative approach to Galois theory by leveraging machine learning techniques to address challenges in understanding polynomial properties and their Galois groups. Combining classical algebraic structures with computational tools opens new avenues for exploring the connections between mathematics and data science.

We have demonstrated the potential of supervised learning to predict Galois groups and polynomial solvability, while unsupervised learning reveals latent structures in polynomial datasets. A comprehensive database of irreducible polynomials with known Galois groups has been compiled, and classical invariants such as discriminants, root differences, and moduli heights have been explored as features for machine learning models. Reduction theories, including Julia and Hermite equivalence, were employed to streamline classification, and the role of polynomial heights in minimal forms and equivalence classes was investigated. The geometric interpretation of polynomial transformations within weighted projective spaces further enhances this framework.

Future work could extend the polynomial database to higher degrees, incorporate multivariable polynomials, and develop novel invariants derived from machine learning. Advanced models, such as graph neural networks, could refine the analysis of root interactions and symmetries, while transfer learning may generalize insights to more complex cases. Automation of reduction methods and interactive visualization tools could make these techniques accessible to a broader audience. Additionally, extending this framework to analyze field extensions and connections with algebraic geometry or physics could broaden its impact.

This work demonstrates the feasibility of integrating machine learning with classical mathematics, offering new tools for algebraists while uncovering deeper theoretical insights. By bridging abstract mathematics and computational science, this approach paves the way for a more interdisciplinary perspective in mathematical research.

\bibliographystyle{ACM-Reference-Format}
\bibliography{references}


\begin{thebibliography}{21}


\ifx \showCODEN    \undefined \def \showCODEN     #1{\unskip}     \fi
\ifx \showDOI      \undefined \def \showDOI       #1{#1}\fi
\ifx \showISBNx    \undefined \def \showISBNx     #1{\unskip}     \fi
\ifx \showISBNxiii \undefined \def \showISBNxiii  #1{\unskip}     \fi
\ifx \showISSN     \undefined \def \showISSN      #1{\unskip}     \fi
\ifx \showLCCN     \undefined \def \showLCCN      #1{\unskip}     \fi
\ifx \shownote     \undefined \def \shownote      #1{#1}          \fi
\ifx \showarticletitle \undefined \def \showarticletitle #1{#1}   \fi
\ifx \showURL      \undefined \def \showURL       {\relax}        \fi
\providecommand\bibfield[2]{#2}
\providecommand\bibinfo[2]{#2}
\providecommand\natexlab[1]{#1}
\providecommand\showeprint[2][]{arXiv:#2}

\bibitem[Berwick(1928)]%
        {berwick-28}
\bibfield{author}{\bibinfo{person}{W.~E.~H. Berwick}.}
  \bibinfo{year}{1928}\natexlab{}.
\newblock \showarticletitle{On {S}oluble {S}extic {E}quations}.
\newblock \bibinfo{journal}{\emph{Proc. London Math. Soc. (2)}}
  \bibinfo{volume}{29}, \bibinfo{number}{1} (\bibinfo{year}{1928}),
  \bibinfo{pages}{1--28}.
\newblock
\showISSN{0024-6115}
\urldef\tempurl%
\url{https://doi.org/10.1112/plms/s2-29.1.1}
\showDOI{\tempurl}


\bibitem[Bhargava and Shankar(2015)]%
        {bhargava-2015}
\bibfield{author}{\bibinfo{person}{Manjul Bhargava} {and} \bibinfo{person}{Arul
  Shankar}.} \bibinfo{year}{2015}\natexlab{}.
\newblock \showarticletitle{Binary quartic forms having bounded invariants, and
  the boundedness of the average rank of elliptic curves}.
\newblock \bibinfo{journal}{\emph{Ann. of Math. (2)}} \bibinfo{volume}{181},
  \bibinfo{number}{1} (\bibinfo{year}{2015}), \bibinfo{pages}{191--242}.
\newblock
\showISSN{0003-486X,1939-8980}
\urldef\tempurl%
\url{https://doi.org/10.4007/annals.2015.181.1.3}
\showDOI{\tempurl}


\bibitem[Bialostocki and Shaska(2005)]%
        {2004-1}
\bibfield{author}{\bibinfo{person}{A. Bialostocki} {and} \bibinfo{person}{T.
  Shaska}.} \bibinfo{year}{2005}\natexlab{}.
\newblock \showarticletitle{Galois groups of prime degree polynomials with
  nonreal roots}.
\newblock In \bibinfo{booktitle}{\emph{Computational aspects of algebraic
  curves}}. \bibinfo{series}{Lecture Notes Ser. Comput.},
  Vol.~\bibinfo{volume}{13}. \bibinfo{publisher}{World Sci. Publ., Hackensack,
  NJ}, \bibinfo{pages}{243--255}.
\newblock
\urldef\tempurl%
\url{https://doi.org/10.1142/9789812701640_0015}
\showURL{%
\tempurl}


\bibitem[Curri(2022)]%
        {curri}
\bibfield{author}{\bibinfo{person}{Elira Curri}.}
  \bibinfo{year}{2022}\natexlab{}.
\newblock \showarticletitle{On the stability of binary forms and their weighted
  heights}.
\newblock \bibinfo{journal}{\emph{Albanian J. Math.}} \bibinfo{volume}{16},
  \bibinfo{number}{1} (\bibinfo{year}{2022}), \bibinfo{pages}{3--23}.
\newblock
\showISSN{1930-1235}


\bibitem[del R\'io and England(2024)]%
        {england-25}
\bibfield{author}{\bibinfo{person}{Tereso del R\'io} {and}
  \bibinfo{person}{Matthew England}.} \bibinfo{year}{2024}\natexlab{}.
\newblock \showarticletitle{Lessons on datasets and paradigms in machine
  learning for symbolic computation: a case study on {CAD}}.
\newblock \bibinfo{journal}{\emph{Math. Comput. Sci.}} \bibinfo{volume}{18},
  \bibinfo{number}{3} (\bibinfo{year}{2024}), \bibinfo{pages}{Paper No. 17,
  27}.
\newblock
\showISSN{1661-8270,1661-8289}
\urldef\tempurl%
\url{https://doi.org/10.1007/s11786-024-00591-0}
\showDOI{\tempurl}


\bibitem[Dolgachev(2003)]%
        {dolgachev}
\bibfield{author}{\bibinfo{person}{Igor Dolgachev}.}
  \bibinfo{year}{2003}\natexlab{}.
\newblock \bibinfo{booktitle}{\emph{Lectures on invariant theory}}.
  \bibinfo{series}{Lond. Math. Soc. Lect. Note Ser.},
  Vol.~\bibinfo{volume}{296}.
\newblock \bibinfo{publisher}{Cambridge: Cambridge University Press}.
\newblock
\showISBNx{0-521-52548-9}
\showISSN{0076-0552}


\bibitem[Hagedorn(2000)]%
        {solv-sextics}
\bibfield{author}{\bibinfo{person}{Thomas~R. Hagedorn}.}
  \bibinfo{year}{2000}\natexlab{}.
\newblock \showarticletitle{General formulas for solving solvable sextic
  equations}.
\newblock \bibinfo{journal}{\emph{J. Algebra}} \bibinfo{volume}{233},
  \bibinfo{number}{2} (\bibinfo{year}{2000}), \bibinfo{pages}{704--757}.
\newblock
\showISSN{0021-8693}
\urldef\tempurl%
\url{https://doi.org/10.1006/jabr.2000.8428}
\showDOI{\tempurl}


\bibitem[King(1996)]%
        {king}
\bibfield{author}{\bibinfo{person}{R.~Bruce King}.}
  \bibinfo{year}{1996}\natexlab{}.
\newblock \bibinfo{booktitle}{\emph{Beyond the quartic equation}}.
\newblock \bibinfo{publisher}{Birkh\"{a}user Boston, Inc., Boston, MA}.
  viii+149 pages.
\newblock
\showISBNx{0-8176-3776-1}


\bibitem[Kotsireas and Shaska(2025)]%
        {2024-06}
\bibfield{author}{\bibinfo{person}{Ilias Kotsireas} {and} \bibinfo{person}{Tony
  Shaska}.} \bibinfo{year}{2025}\natexlab{}.
\newblock \showarticletitle{A machine learning approach of {J}ulia reduction}.
  In \bibinfo{booktitle}{\emph{(submitted)}}. ISSAC.
\newblock
\urldef\tempurl%
\url{https://www.risat.org/pdf/2024-06.pdf}
\showURL{%
\tempurl}


\bibitem[Lample and Charton(2019)]%
        {lample-charton}
\bibfield{author}{\bibinfo{person}{Guillaume Lample} {and}
  \bibinfo{person}{Fran{\c c}ois Charton}.} \bibinfo{year}{2019}\natexlab{}.
\newblock \showarticletitle{Deep Learning for Symbolic Mathematics}.
\newblock  (\bibinfo{year}{2019}).
\newblock
\showeprint[arxiv]{1912.01412}~[cs.SC]
\urldef\tempurl%
\url{https://arxiv.org/abs/1912.01412}
\showURL{%
\tempurl}


\bibitem[Noorbakhsh et~al\mbox{.}(2023)]%
        {nsn-2}
\bibfield{author}{\bibinfo{person}{Kimia Noorbakhsh}, \bibinfo{person}{Modar
  Sulaiman}, \bibinfo{person}{Mahdi Sharifi}, \bibinfo{person}{Kallol Roy},
  {and} \bibinfo{person}{Pooyan Jamshidi}.} \bibinfo{year}{2023}\natexlab{}.
\newblock \showarticletitle{Pretrained Language Models are Symbolic Mathematics
  Solvers too!}
\newblock  (\bibinfo{year}{2023}).
\newblock
\showeprint[arxiv]{2110.03501}~[stat.ML]
\urldef\tempurl%
\url{https://arxiv.org/abs/2110.03501}
\showURL{%
\tempurl}


\bibitem[Pickering et~al\mbox{.}(2024)]%
        {england}
\bibfield{author}{\bibinfo{person}{Lynn Pickering}, \bibinfo{person}{Tereso del
  R\'io~Almajano}, \bibinfo{person}{Matthew England}, {and}
  \bibinfo{person}{Kelly Cohen}.} \bibinfo{year}{2024}\natexlab{}.
\newblock \showarticletitle{Explainable {AI} insights for symbolic computation:
  a case study on selecting the variable ordering for cylindrical algebraic
  decomposition}.
\newblock \bibinfo{journal}{\emph{J. Symbolic Comput.}}  \bibinfo{volume}{123}
  (\bibinfo{year}{2024}), \bibinfo{pages}{Paper No. 102276, 24}.
\newblock
\showISSN{0747-7171,1095-855X}
\urldef\tempurl%
\url{https://doi.org/10.1016/j.jsc.2023.102276}
\showDOI{\tempurl}


\bibitem[Schur(1968)]%
        {schur}
\bibfield{author}{\bibinfo{person}{I. Schur}.} \bibinfo{year}{1968}\natexlab{}.
\newblock \bibinfo{booktitle}{\emph{Vorlesungen {\"u}ber
  {Invariantentheorie}}}. \bibinfo{series}{Grundlehren Math. Wiss.},
  Vol.~\bibinfo{volume}{143}.
\newblock \bibinfo{publisher}{Springer, Cham}.
\newblock
\showISSN{0072-7830}


\bibitem[Serre(1992)]%
        {serre}
\bibfield{author}{\bibinfo{person}{Jean-Pierre Serre}.}
  \bibinfo{year}{1992}\natexlab{}.
\newblock \bibinfo{booktitle}{\emph{Topics in {G}alois theory}}.
  \bibinfo{series}{Research Notes in Mathematics}, Vol.~\bibinfo{volume}{1}.
\newblock \bibinfo{publisher}{Jones and Bartlett Publishers, Boston, MA}.
  xvi+117 pages.
\newblock
\showISBNx{0-86720-210-6}
\newblock
\shownote{Lecture notes prepared by Henri Damon [Henri Darmon], With a foreword
  by Darmon and the author}.


\bibitem[Shaska and Shaska(2024a)]%
        {2024-07}
\bibfield{author}{\bibinfo{person}{Elira Shaska} {and} \bibinfo{person}{Tony
  Shaska}.} \bibinfo{year}{2024}\natexlab{a}.
\newblock \showarticletitle{Irreducible sextics, invariants, and their Galois
  groups}.
\newblock \bibinfo{howpublished}{https://www.risat.org/pdf/2024-07.pdf}.
\newblock \bibinfo{journal}{\emph{RISAT preprints}} (\bibinfo{date}{12}
  \bibinfo{year}{2024}).
\newblock
\showeprint[risat]{https://www.risat.org/pdf/2024-07.pdf}~[math.AG]
\urldef\tempurl%
\url{https://www.risat.org/pdf/2024-07.pdf}
\showURL{%
\tempurl}


\bibitem[Shaska and Shaska(2024b)]%
        {2024-03}
\bibfield{author}{\bibinfo{person}{Elira Shaska} {and} \bibinfo{person}{Tony
  Shaska}.} \bibinfo{year}{2024}\natexlab{b}.
\newblock \showarticletitle{Machine learning for moduli space of genus two
  curves and an application to isogeny based cryptography}.
\newblock  (\bibinfo{year}{2024}).
\newblock
\showeprint[arxiv]{2403.17250}~[math.AG]
\urldef\tempurl%
\url{https://arxiv.org/abs/2403.17250}
\showURL{%
\tempurl}


\bibitem[Shaska and Shaska(2024c)]%
        {2024-05}
\bibfield{author}{\bibinfo{person}{Elira Shaska} {and} \bibinfo{person}{Tony
  Shaska}.} \bibinfo{year}{2024}\natexlab{c}.
\newblock \showarticletitle{Polynomials, Galois groups, and Deep Learning}.
\newblock \bibinfo{howpublished}{https://www.risat.org/pdf/2024-05.pdf}.
\newblock \bibinfo{journal}{\emph{RISAT preprints}} (\bibinfo{date}{12}
  \bibinfo{year}{2024}).
\newblock
\showeprint[risat]{https://www.risat.org/pdf/2024-05.pdf}~[math.AG]
\urldef\tempurl%
\url{https://www.risat.org/pdf/2024-05.pdf}
\showURL{%
\tempurl}


\bibitem[Shaska and Shaska(2025)]%
        {galois-web}
\bibfield{author}{\bibinfo{person}{Elira Shaska} {and} \bibinfo{person}{Tony
  Shaska}.} \bibinfo{year}{2025}\natexlab{}.
\newblock \bibinfo{title}{Galois Theory: A database approach}.
\newblock
\newblock
\urldef\tempurl%
\url{https://www.risat.org/galois.html}
\showURL{%
\tempurl}


\bibitem[Shaska(2022)]%
        {2020-1}
\bibfield{author}{\bibinfo{person}{T. Shaska}.}
  \bibinfo{year}{2022}\natexlab{}.
\newblock \showarticletitle{Reduction of superelliptic {R}iemann surfaces}.
\newblock In \bibinfo{booktitle}{\emph{Automorphisms of {R}iemann surfaces,
  subgroups of mapping class groups and related topics}}.
  \bibinfo{series}{Contemp. Math.}, Vol.~\bibinfo{volume}{776}.
  \bibinfo{publisher}{Amer. Math. Soc., [Providence], RI},
  \bibinfo{pages}{227--247}.
\newblock
\urldef\tempurl%
\url{https://doi.org/10.1090/conm/776/15614}
\showDOI{\tempurl}


\bibitem[Shaska and Beshaj(2015)]%
        {2014-1}
\bibfield{author}{\bibinfo{person}{T. Shaska} {and} \bibinfo{person}{L.
  Beshaj}.} \bibinfo{year}{2015}\natexlab{}.
\newblock \showarticletitle{Heights on algebraic curves}. In
  \bibinfo{booktitle}{\emph{Advances on superelliptic curves and their
  applications}} \emph{(\bibinfo{series}{NATO Sci. Peace Secur. Ser. D Inf.
  Commun. Secur.}, Vol.~\bibinfo{volume}{41})}. \bibinfo{publisher}{IOS Press},
  \bibinfo{address}{Amsterdam}, \bibinfo{pages}{137--175}.
\newblock
\showISBNx{978-1-61499-519-7; 978-1-61499-520-3}


\bibitem[V\"olklein(1996)]%
        {helmut-book}
\bibfield{author}{\bibinfo{person}{Helmut V\"olklein}.}
  \bibinfo{year}{1996}\natexlab{}.
\newblock \bibinfo{booktitle}{\emph{Groups as {G}alois groups}}.
  \bibinfo{series}{Cambridge Studies in Advanced Mathematics},
  Vol.~\bibinfo{volume}{53}.
\newblock \bibinfo{publisher}{Cambridge University Press, Cambridge}. xviii+248
  pages.
\newblock
\showISBNx{0-521-56280-5}
\urldef\tempurl%
\url{https://doi.org/10.1017/CBO9780511471117}
\showDOI{\tempurl}
\newblock
\shownote{An introduction}.


\end{thebibliography}

\newpage

\appendix

\section{Results and Implementation}\label{app:quintics}
Here we present all results for cubics, quartics, and quintics from our data. 

\subsection{Cubics}

\begin{figure}[h!] 
   \centering
   \includegraphics[width=2in]{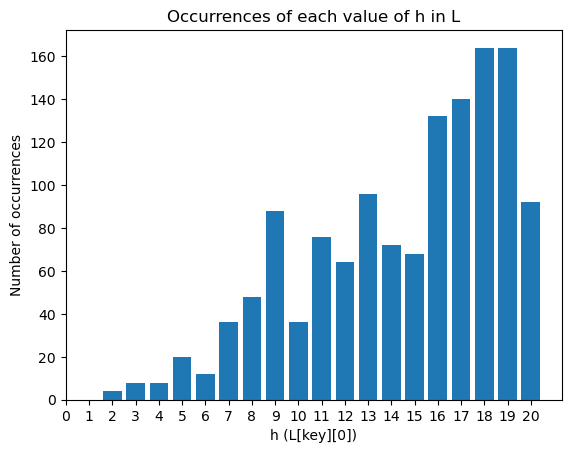} 
   \caption{Distribution of cubics with Galois group $C_3$.}
   \label{fig:deg3}   
\end{figure}


\begin{small}
\begin{table}[h!]
\begin{center}
\begin{tabular}{ | c | c | c ||   c | c | c || c | c | c |  }
\hline
 \# &   $f$    &  ${\Delta}$ & 	\# &   $f$    &  $\Delta$     \\
 \hline
1 	& 	( 1, 3, -4, 1 )      & $7^2$ 	&	 21	&	(1, -4, 1, 1) 	&	$13^2$	  \\
2 	&	( -1, -4, -3, 1 )    & $7^2$  	&	22	&	(-5, -3, 2, 1)	&	$13^2$	  \\
3	&	( 1, -1, -2, 1)     & $7^2$  	&	23	&	(-1, 1, 4, 1)	&	 $13^2$	 \\ 
4	&	(1, -2, -1, 1)     & $7^2$  	&	24	&	(-5, 4, 5, 1) 	&	$13^2$	  \\ 
5	&	(-1, -2, 1, 1)     & $7^2$  	&	25	&	(-1, -5, -4, 5)	&	$13^2$	  \\ 
6	&	(-1, -1, 2, 1)     & $7^2$  	&	26	&	(1, -2, -3, 5)	&	$13^2$	  \\
7	&	(1, -4, 3, 1)      & $7^2$  	&	27	&	(-1, -2, 3, 5)	&	 $13^2$   \\
8	&	(-1, 3, 4, 1)      & $7^2$  	&	28	&	(1, -5, 4, 5)	&	 $13^2$    \\
9	&	(1, 0, -3, 1)      & $3^4$  	&	29	&	  (1, 2, -5, 1)      & $19^2$	 \\
10	&	(3, 0, -3, 1)      & $3^4$  	&	30	&	(-1, -5, -2, 1)    & $19^2$	 \\
11	&	(-1, -3, 0, 1)     & $3^4$  	&	31	&	(1, -5, 2, 1)      & $19^2$	  \\
12	&	(1, -3, 0, 1)      & $3^4$   	&	32	&	(-1, 2, 5, 1)      & $19^2$	  \\
13	&	(-3, 0, 3, 1)      & $3^4$   	&	33	&	(2, -1, -5, 2)     & $31^2$	 \\ 
14	&	(-1, 0, 3, 1)      & $3^4$   	&	32 	&	(2, -5, -1, 2)     & $31^2$	  \\
15	&	(-1, -3, 0, 3) 	& 		&	 35	&	(-2, -5, 1, 2)     & $31^2$	  \\
16	&	(1, -3, 0, 3)	&  		&	36	&	(-2, -1, 5, 2)     & $31^2$	  \\
17	&	(5, 4, -5, 1)	&		&  	37	&	(3, -4, -5, 3)     &  $61^2$	 \\
18	&	(1, 1, -4, 1) 	&		&  	38	&	(3, -5, -4, 3)     &  $61^2$	  \\
19	&	(5, -3, -2, 1) 	&		&  	39	&	(-3, -5, 4, 3)     &  $61^2$	  \\
20	&	(-1, -4, -1, 1)	&		&  	40	&	(-3, -4, 5, 3)     &  $61^2$	  \\
\hline
\end{tabular}
\end{center}
\caption{Cubics of height $\leq 5$ and Galois group $C_3$}
\label{tab:deg-3}
\end{table}
\end{small}

\vspace{-0.8cm}

\begin{figure}[h!] 
   \centering
   \includegraphics[width=2in]{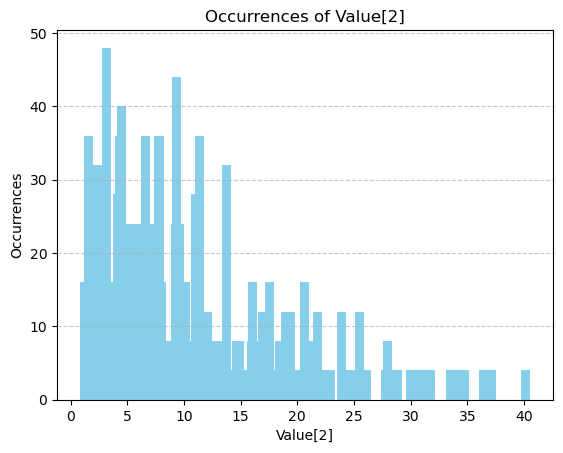} 
   \caption{Occurrences for cubics versus the invariants}
   \label{fig:deg3-2}
\end{figure}

\subsection{Quartics}

\begin{figure}[h!] 
   \centering
   \includegraphics[width=2in]{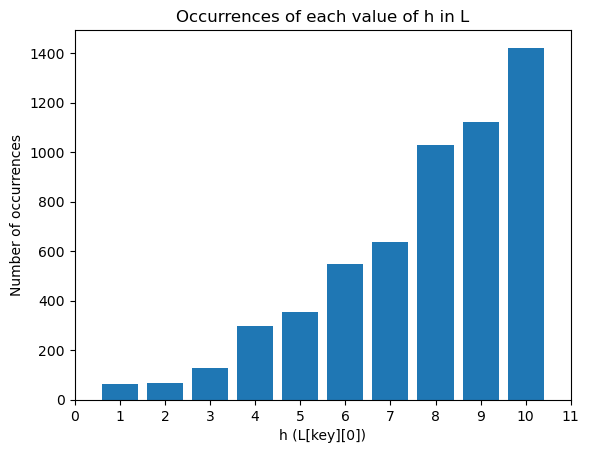} 
   \caption{Distribution of quartics with $\Gal (f) \not\cong S_4$.}
   \label{fig:deg4}   
\end{figure}


\begin{figure}[h!] 
   \centering
   \includegraphics[width=2in]{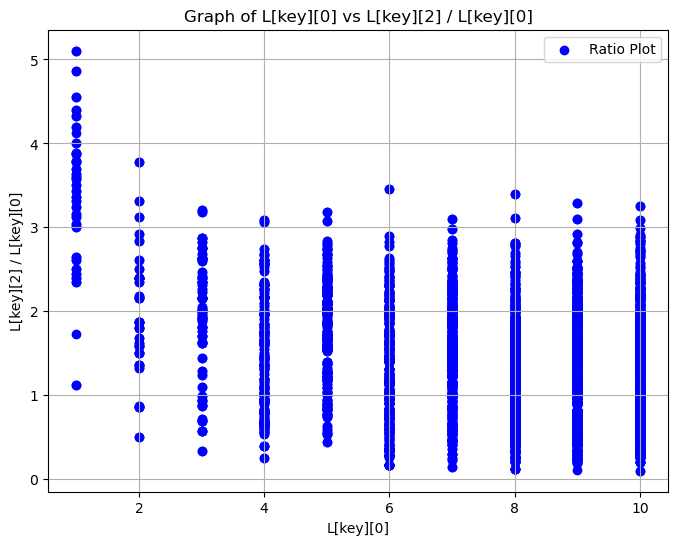} 
      \caption{The ratio of weighted height with naive height}
   \label{fig:deg4-2}
\end{figure}

\subsection{Quintics} \label{app:quintics}

\begin{figure}[h!] 
   \centering
   \includegraphics[width=2in]{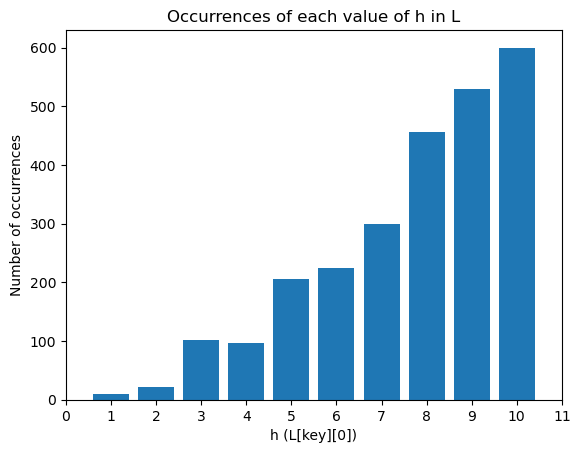} 
   \caption{Distribution of quintics with $\Gal (f) \not\equiv S_5$.}
   \label{fig:deg5}
\end{figure}


\begin{figure}[h!] 
   \centering
   \includegraphics[width=2in]{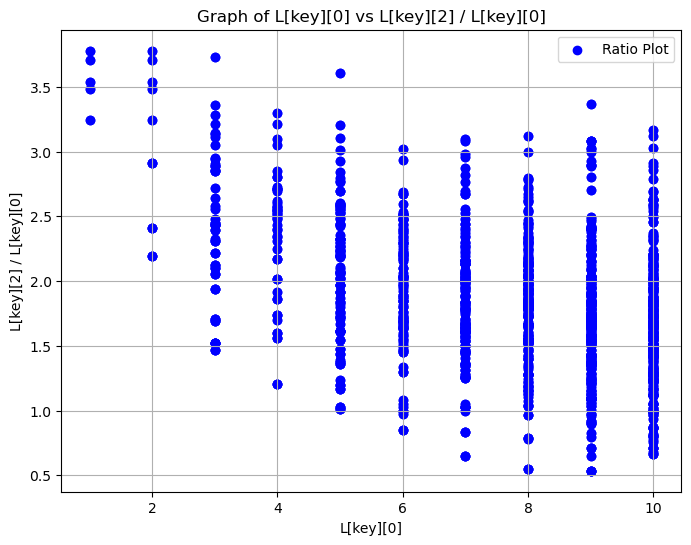} 
      \caption{The ratio of weighted height with naive height}
   \label{fig:deg5-2}
\end{figure}

 
Next is  presented  the  Python   implementation of the symbolic  layers  for irreducible quintics: 

\begin{lstlisting}[caption={Python implementation of the \texttt{sig\_layer} function.}, label={lst:siglayer}]
from sympy import symbols, Poly, factor_list
def sig_layer(p):
    x = symbols('x')
    f = sum(a * x**i for i, a in enumerate(p))
    signature = [5]
    primes = [2, 3, 5, 7]
    for prime in primes:
        poly_mod = Poly(f, x, modulus=prime)
        factors = factor_list(poly_mod)[1]          
        for factor_poly, multiplicity in factors:
            degree = factor_poly.degree()
            if degree > 1 and degree not in signature:   
                signature.append(degree)
    return signature
\end{lstlisting}

\begin{lstlisting}[language=Python, caption=Real Root Counting Algorithm]
from sympy import symbols, diff, Poly, sign

def sturm_sequence(P, x):
    P = Poly(P, x)   
    sequence = [P, P.diff(x)]   
    while True:
        remainder = -sequence[-2].rem(sequence[-1])   
        if remainder.is_zero:
            break
        sequence.append(remainder)
    return sequence

def count_sign_changes(sequence, value):
    evaluations = []
    for poly in sequence:
        eval_value = poly.eval(value)
        if eval_value == 0:
            evaluations.append(0)   
        else:
            evaluations.append(sign(eval_value))
    evaluations = [s for i, s in enumerate(evaluations) if i == 0 or s != evaluations[i - 1]]
    return len(evaluations) - 1

def real_root_count(P, x, interval=(-1e10, 1e10)):
    a, b = interval
    P = Poly(P.expand(), x)  
    sturm_seq = sturm_sequence(P, x)
    sign_changes_a=count_sign_changes(sturm_seq,a)
    sign_changes_b=count_sign_changes(sturm_seq,b)
    return sign_changes_a - sign_changes_b
\end{lstlisting}

\end{document}